\documentclass[letterpaper, 10 pt, conference]{ieeeconf}  

\usepackage{amsmath,amssymb,mathtools,bm,graphicx,booktabs,multirow}
\usepackage{array}
\usepackage{cite}
\usepackage{url}
\usepackage{hyperref}
\usepackage{microtype}
\newtheorem{prop}{Proposition}
\newtheorem{rmk}{Remark}
\newcommand{\R}{\mathbb{R}}
\newcommand{\Sset}{\mathcal{S}}
\newcommand{\Dset}{\mathcal{D}}
\newcommand{\Xset}{\mathcal{X}}
\newcommand{\Yset}{\mathcal{Y}}
\newcommand{\Qcert}{\mathcal{Q}_{\mathrm{cert}}}
\newcommand{\relu}{\operatorname{ReLU}}
\newcommand{\sat}{\operatorname{sat}}
\allowdisplaybreaks

\usepackage{MnSymbol}

\usepackage[scaled]{beramono}
\usepackage[T1]{fontenc}

\IEEEoverridecommandlockouts                              
\title{Quadratic Characterizations for Reachability Analysis of Neural Networks}

\author{Elias Khalife, Mazen Farhood, and Pierre-Loic Garoche
\thanks{E. Khalife and P.-L. Garoche are with the Federation ENAC ISAE-SUPAERO ONERA, Universite de Toulouse, Toulouse, France 
(e-mail: \{elias.khalife, pierre-loic.garoche\}@enac.fr). M. Farhood is with the Kevin T. Crofton Department of Aerospace and Ocean Engineering, Virginia Tech, Blacksburg, VA 24061, USA (e-mail: farhood@vt.edu).}
\thanks{}
}

\begin{document}

\maketitle
\thispagestyle{empty}
\pagestyle{empty}

\begin{abstract}
Quadratic constraints (QCs) are widely used to characterize nonlinearities and uncertainties, but generic analytical characterizations can be conservative on bounded domains. This paper develops a framework for constructing verified quadratic characterizations of scalar relations in the two-dimensional real plane. Candidate quadratic inequalities are locally generated by solving convex quadratic programs using samples from the relation and exterior sample points. They are then verified globally using sum-of-squares certificates over an exact semialgebraic description or, in the case of nonpolynomial relations, over relaxed polynomial descriptions. The resulting verified constraints define a sound overapproximation of the scalar relations over the considered domains. These constraints are directly compatible with existing analysis frameworks based on QCs and pointwise integral quadratic constraints (IQCs) for static nonlinearities and uncertainties, and they can also be embedded in QC-based semidefinite programs for reachability and safety analysis of feedforward neural networks. For smooth activations such as $\tanh$, the method yields domain-dependent quadratic characterizations that constitute an alternative to generic sector- or slope-based descriptions. For ReLU networks, we give methods to reduce conservatism in QC-based reachability analysis of feedforward networks by exploiting dependencies between neurons and tighter local bounds. Numerical examples demonstrate improved reachability results for smooth activations, reduced conservatism for ReLU networks, and applicability beyond neural networks through an example involving saturation.
\end{abstract}

\section{Introduction}
In systems and control, quadratic constraints (QCs) are commonly used to characterize static input-output relations, including sector-bounded, slope-restricted, monotone, and repeated nonlinearities, as well as other static uncertainties \cite{megretski1997,veenman2016,khalife2023}. QCs are also used in neural network verification and reachability analysis problems to characterize activation functions \cite{fazlyab2022,hu2020}. 
Standard QC descriptions used in analysis are often derived analytically from broad structural properties of the underlying nonlinearity, such as sector, slope, monotonicity, or repetition conditions \cite{veenman2016,fry2021robustness}. These descriptions are convenient and often effective, but they are not, in general, tailored to a given static relation over a compact domain, which can lead to conservative analysis results. In this work, we seek a finite family of verified pointwise quadratic inequalities that defines a sound overapproximation of the relevant portion of the graph for a given static relation over a compact domain. This is motivated by applications in which only a bounded subset of the set of input-output pairs defining the relation is relevant to the subsequent analysis. For example, a sector overapproximation of saturation may be overly conservative when restricted to a specific domain of interest. A similar issue arises in neural network reachability analysis, where admissible preactivation ranges are determined by the input set and the network parameters, so only bounded portions of the activation graphs need to be characterized.

This paper gives an approach for generating such QCs, which entails first generating candidate quadratic inequalities based on judiciously chosen sample points by solving convex quadratic programs, followed by verifying these inequalities using sum-of-squares certificates over an exact semialgebraic description of the relation or, in the nonpolynomial case, over a relaxed polynomial description \cite{putinar1993,parrilo2005}. The resulting family of verified quadratic inequalities defines a sound outer approximation of the relation over the domain of interest and can be integrated into established QC-based and some IQC-based analysis methods \cite{he2022,khalife2023}. In this paper, we work out this integration for QC-based semidefinite programs arising in reachability and safety analysis problems of feedforward neural networks \cite{fazlyab2022,hu2020}. For smooth activations such as $\tanh$, the proposed approach is used to generate the associated QCs, which can replace or complement generic analytical characterizations on the relevant bounded domains. For ReLU networks, the scalar ReLU graph already admits an exact quadratic characterization \cite{fazlyab2022}. To reduce the conservatism in this case, we focus on combining this exact characterization with blockwise coupling of repeated ReLU units \cite{noori2025} and tightening of the neurons' pre- and post-activation bounds. Several examples are provided to demonstrate the generality and utility of the proposed approach in neural network and uncertain system analysis.

The remainder of the paper is organized as follows. Section~\ref{sec:prelim} presents the notation and problem setup. Section~\ref{sec:qc_construction} develops the sample-based candidate-generation and SOS-verification procedure. Section~\ref{sec:reachability_application} incorporates the verified quadratic characterizations into QC-based formulations for neural network reachability and safety analysis. Section~\ref{sec:relu_tightening} presents the ReLU-specific improvements. Section~\ref{sec:examples} demonstrates the developed methods through examples, and Section~\ref{sec:conclusion} concludes the paper.

\section{Preliminaries}\label{sec:prelim}
\subsection{Notation}\label{sec:notation}
The sets of real numbers, strictly positive integers, $n$-dimensional real vectors, $n\times m$ real matrices, and $n\times n$ real symmetric matrices are denoted by $\mathbb{R}$, $\mathbb{Z}_+$, $\mathbb{R}^n$, $\mathbb{R}^{n\times m}$, and $\mathbb{S}^n$, respectively. The symbol $I_n$ denotes the $n\times n$ identity matrix, and $\mathbf{1}$ denotes the vector of all ones. Subscripts indicating dimensions are omitted when clear from the context. For a matrix or vector $X$, $X^\top$ denotes its transpose. For a finite index set $\mathcal{I}$, $|\mathcal{I}|$ denotes its cardinality. For a symmetric matrix $X\in\mathbb{S}^n$, we write $X\succeq 0$ and $X\preceq 0$ to mean that $X$ is positive semidefinite and negative semidefinite, respectively.

For $x \in \mathbb{R}^n$, $\|x\|_2$ and $\|x\|_\infty$ denote the Euclidean and infinity norms, respectively. For a polynomial variable $z$, $\Sigma[z]$ denotes the cone of sum-of-squares polynomials in $z$.

\subsection{Problem Setup}\label{sec:problem_setup}
Let $\Sset \subset \mathbb{R}^2$ denote the set of admissible input-output pairs $z=(x,y)$ satisfying a prescribed relation, and let $\Dset\subset\mathbb{R}$ be a compact input domain. We denote by $\Sset_{\Dset}:=\{(x,y)\in\Sset \mid x\in\Dset\}$ the restriction of $\Sset$ to $\Dset$. The relation may be specified by algebraic equalities and inequalities, logical conditions, piecewise functions, or any finite combination thereof. Our objective is to obtain a finite family of quadratic functions that are nonnegative on $\Sset_{\Dset}$.

To this end, define the quadratic monomial basis
\begin{align*}
    \phi(z):=\begin{bmatrix} x^2 & y^2 & xy & x & y & 1\end{bmatrix}^{\top}.
\end{align*}
For $i=1,\dots,N_q$, we consider the quadratic functions
\begin{equation*}
q_i(z)=
\begin{bmatrix}
z^\top & 1
\end{bmatrix}
Q_i
\begin{bmatrix}
z \\ 1
\end{bmatrix}
=c_i^\top \phi(z),
\end{equation*}
where $Q_i \in \mathbb{S}^3$, $c_i \in \mathbb{R}^6$, and $N_q\in\mathbb{Z}_+$.
The family $\{q_i\}_{i=1}^{N_q}$ is a quadratic characterization of $\Sset_{\Dset}$ if
\begin{equation}
q_i(z) \ge 0,
\quad
\forall\, z \in \Sset_{\Dset},
\quad
i=1,\dots,N_q.
\end{equation}
The associated set
\begin{equation*}
\widehat{\Sset}_{N_q}
:=
\left\{
(x,y) \in \Dset\times\R \mid q_i(x,y) \ge 0,\; i=1,\dots,N_q
\right\}
\end{equation*}
is a sound overapproximation of $\Sset_{\Dset}$ since $\Sset_{\Dset} \subseteq \widehat{\Sset}_{N_q}$.

Increasing the number of constraints $N_q$ can improve the tightness of the overapproximation at the expense of added computational complexity in ensuing analysis. Therefore, the problem is to compute a verified quadratic characterization that balances the tightness of the overapproximation with the computational cost of the analysis.

\section{Data-Driven Candidate Generation and SOS Verification}\label{sec:qc_construction}

The proposed approach consists of two stages. In the first stage, candidate quadratic constraints are generated from sampled data. In the second stage, each candidate is verified by solving SOS feasibility problems.

\subsection{Local decomposition, sampling, and candidate generation}

Rather than constructing candidate quadratic functions on the entire $\Sset_{\Dset}$, we generate them on local restrictions of the relation. Let $\mathcal{U}_k\subseteq\Dset$ be subdomains satisfying $\Dset=\bigcup_{k=1}^{N_d}\mathcal{U}_k$ for some $N_d\in\mathbb{Z}_+$, and define $\Sset_k:=\Sset_{\mathcal{U}_k}=\{(x,y)\in\Sset \mid x\in\mathcal{U}_k\}$. The subdomains do not have to be disjoint and are chosen to isolate local geometric features, such as smooth branches, corners, switching regions, or junctions induced by piecewise or logical descriptions.

\noindent For each subdomain $\mathcal{U}_k$, we use three classes of~sample~points:
\begin{align*}
z_{k,j}^{\mathrm{loc}} &\in \Sset_k,
\qquad j=1,\dots,n_k^{\mathrm{loc}}, \\
z_j^{\mathrm{g}} &\in \Sset_{\Dset},
\qquad j=1,\dots,n^{\mathrm{g}}, \\
z_{k,j}^{\mathrm{ext}} &\notin \Sset,
\qquad j=1,\dots,m_k,
\end{align*}
for some strictly positive integers $n_k^{\mathrm{loc}}$, $n^{\mathrm{g}}$, and $m_k$, where $k\in\{1,\ldots,N_d\}$. Here, $z_{k,j}^{\mathrm{loc}}$ are used to shape the candidate function on the local restriction $\Sset_k$, while $z_j^{\mathrm{g}}$ help enforce the required nonnegativity condition on $\Sset_{\Dset}$. The exterior samples $z_{k,j}^{\mathrm{ext}}$ are points outside $\Sset$ where a negativity condition is imposed, typically chosen depending on the desired geometry of the resulting quadratic constraint. In practice, local samples are often placed near boundaries, transitions, corners, or other geometrically informative regions.

For each restriction $\Sset_k$, we compute a candidate function
\begin{equation}
q_k^{\mathrm{cand}}(z)=c_k^\top\phi(z),
\qquad c_k\in\mathbb{R}^6,
\end{equation}
by solving a convex quadratic program:
\begin{equation}
\label{eq:local_qp}
\begin{aligned}
\min_{c_k,\xi_k^{\mathrm{loc}},\,\xi_k^{\mathrm{g}},\eta_k}\quad
&
\frac{\rho}{2}\|c_k\|_2^2
+\lambda_{\mathrm{loc}} \mathbf{1}^\top \xi_k^{\mathrm{loc}}
+\lambda_{\mathrm{g}} \mathbf{1}^\top \xi_k^{\mathrm{g}}
+\lambda_{\mathrm{ext}} \mathbf{1}^\top \eta_k
\\
\text{s.t.}\quad
&
c_k^\top \phi\!\left(z_{k,j}^{\mathrm{loc}}\right)
\ge
\bar{\gamma}-\xi_{k,j}^{\mathrm{loc}},
\qquad
j=1,\dots,n_k^{\mathrm{loc}},
\\
&
c_k^\top \phi\!\left(z_j^{\mathrm{g}}\right)
\ge
\bar{\gamma}-\xi_{k,j}^{\mathrm{g}},
\qquad
j=1,\dots,n^{\mathrm{g}},
\\
&
c_k^\top \phi\!\left(z_{k,j}^{\mathrm{ext}}\right)
\le
-\bar{\gamma}+\eta_{k,j},
\qquad
j=1,\dots,m_k,\\
&
\xi_k^{\mathrm{loc}}\ge 0,\qquad
\xi_k^{\mathrm{g}}\ge 0,\qquad
\eta_k\ge 0,
\end{aligned}
\end{equation}
where $\bar{\gamma}>0$ is a prescribed margin, $\lambda_{\mathrm{loc}},\lambda_{\mathrm{g}},\lambda_{\mathrm{ext}}>0$ are user-selected weights, $\rho>0$ is a regularization parameter that penalizes the magnitude of the coefficient vector, and 
$\xi_k^{\mathrm{loc}} \in \mathbb{R}^{n_k^{\mathrm{loc}}}$,
$\xi_k^{\mathrm{g}} \in \mathbb{R}^{n^{\mathrm{g}}}$, and
$\eta_k \in \mathbb{R}^{m_k}$ are nonnegative vectors included to allow potential violations at certain sample points. The size of problem \eqref{eq:local_qp} scales linearly with $(n_k^{\mathrm{loc}}+n^{\mathrm{g}}+m_k)$.

Multiple candidate quadratic functions may be computed for a local restriction $\Sset_k$ depending on the local geometry of the relation and on whether the desired QCs are intended to lead to tighter lower or upper bounds. For instance, one may compute separate candidate functions to specify the lower- and upper-bounding curves for the overapproximation of $\mathcal{S}_k$, with the sample sets selected accordingly. However, regardless of the sampling strategy, solving \eqref{eq:local_qp} only enforces the prescribed inequalities on finitely many samples and does not guarantee that $q_k^{\mathrm{cand}}(z)\ge 0$ for all $z\in\Sset_{\Dset}$. A separate verification step is required to ensure global validity on $\Sset_{\Dset}$.

\subsection{SOS-based verification}
\subsubsection{Piecewise-polynomial relations}
Suppose that the set $\mathcal{S}_{\Dset}$ is defined by a piecewise-polynomial relation, i.e., $\mathcal{S}_{\Dset}=\{(x,y)\in\Dset\times\mathbb{R}\,|\, P(x,y)=0\}$, where $P(\cdot,\cdot)$ is a piecewise-polynomial function. In this case, $\mathcal{S}_\Dset$ can be represented exactly by a semialgebraic set or a union of semialgebraic sets. 
Assume $\Sset_{\Dset}$ is expressed as a union of $N_s$ semialgebraic sets for some $N_s\in\mathbb{Z}_+$, i.e.,
\begin{equation*}
    \Sset_{\Dset}=\bigcup_{i=1}^{N_s}\Sset^{(i)},
\end{equation*}
where $\Sset^{(i)}=\left\{z \in \mathbb{R}^2 \,\mid\, g_j^{(i)}(z) \ge 0,\ j=1,\dots,m_i\right\}$ and $i=1,\dots,N_s$. A candidate quadratic $q_k^{\mathrm{cand}}$ is verified by solving the SOS feasibility problem on each subset $\Sset^{(i)}$. Specifically, for each $i=1,\dots,N_s$, we solve the SOS feasibility problem
\begin{equation}
\label{eq:sos_union_feas}
\begin{aligned}
\text{find}\quad
&
\sigma_{1,k}^{(i)},\dots,\sigma_{m_i,k}^{(i)}
\\
\text{s.t.}\quad
&
\sigma_{j,k}^{(i)} \in \Sigma[z],
\qquad
j=1,\dots,m_i,
\\
&
q_k^{\mathrm{cand}}(z)
-
\sum_{j=1}^{m_i}\sigma_{j,k}^{(i)}(z)\,g_j^{(i)}(z)
\in \Sigma[z].
\end{aligned}
\end{equation}
If \eqref{eq:sos_union_feas} is feasible for every $i\in\{1,\dots,N_s\}$, then $q_k^{\mathrm{cand}}(z)\ge 0$ for all $z \in \Sset_{\Dset}$.

Let $\Qcert$ denote the set of candidate quadratic functions that pass the SOS verification step, and let $N_q=|\Qcert|$. Then, $\Qcert$ defines a verified quadratic characterization of $\Sset_{\Dset}$, and the associated set $\hat{\mathcal{S}}_{N_q}$ satisfies $\mathcal{S}_D \subseteq \hat{\mathcal{S}}_{N_q}$.
If $\Sset$ has a known symmetry, candidate generation and verification can be restricted to a representative subset of the domain, and the remaining candidates can be obtained by symmetry.

\begin{rmk}\label{rmk:domain}
The domain used for candidate generation and the domain used for verification do not need to be identical. In particular, candidates may be generated on $\Dset$ and then verified on a larger domain $\widetilde{\Dset} \supseteq \Dset$.
\end{rmk}

\begin{rmk}\label{rmk:alt_verification}
As an alternative to SOS verification, the soundness of the overapproximation can also be established by SMT-based verification \cite{barrett2018satisfiability}, which can account for floating-point round-off errors (see \cite{khalife2023nfm,khalife2025nfm}), and by machine-checked proofs in interactive theorem provers such as Rocq~\cite{bertot2013interactive}.
\end{rmk}

\subsubsection{Nonpolynomial relations}

When the relation defining $\Sset_{\mathcal{D}}$ is not piecewise-polynomial, the verification step is handled differently. The candidate generation step remains data-driven and can involve samples from the exact relation. However, SOS verification requires representing the set $\Sset_{\Dset}$ using polynomial constraints. Therefore, a polynomial approximation is introduced in the verification step, along with a bound on the approximation error.

Consider the set
\begin{equation*}
\Sset_{\Dset}:= \left\{(x,y)\in\mathbb{R}^2 \mid y=\varphi(x),\ x\in\Dset\right\},
\end{equation*}
where $\varphi:\mathbb{R}\to\mathbb{R}$ is a nonpolynomial function.
Let $p$ be a polynomial approximation of $\varphi$ on $\Dset$, and suppose that an error bound $\varepsilon\ge 0$ is available such that
\begin{equation*}
|\varphi(x)-p(x)|\le \varepsilon
\ \ \mbox{for all} \ x\in\Dset.
\end{equation*}
We then define the relaxed verification set
\begin{equation*}
\widetilde{\Sset}_{p,\varepsilon,\Dset} :=
\left\{(x,y)\in\mathbb{R}^2 \mid x\in\Dset,\; -\varepsilon \le y-p(x)\le \varepsilon \right\}.
\end{equation*}
Since $\Sset_{\Dset}\subseteq\widetilde{\Sset}_{p,\varepsilon,\Dset}$, the verification of a candidate quadratic function on $\widetilde{\Sset}_{p,\varepsilon,\Dset}$ carries over to $\Sset_{\Dset}$. The error bound may be derived analytically, for example, from Taylor remainder estimates or from the remainder of polynomial interpolation at Chebyshev nodes~\cite{mason2002chebyshev}.

The approximation is not necessarily constructed over the entire domain $\Dset$ at once. If $\Dset=\bigcup_{i=1}^{N_e}\Dset_i$ for some $N_e\in\mathbb{Z}_+$, one may instead use polynomial approximations $p_i$ with error bounds $\varepsilon_i$ on the subsets $\Dset_i$ and verify the candidate on the corresponding relaxed verification sets $\widetilde{\Sset}_{p_i,\varepsilon_i,\Dset_i}$. Verification on all such sets implies verification on $\Sset_{\Dset}$. This approach is used in the $\tanh$ example in Section~\ref{sec:examples}.

\section{Application to Feedforward Neural Network Reachability Analysis}\label{sec:reachability_application}

In this section, we show how the verified quadratic functions constructed in Section~\ref{sec:qc_construction} can be incorporated into an existing QC-based semidefinite program (SDP) formulation for reachability analysis and safety verification of feedforward neural networks, such as the framework in \cite{fazlyab2022}.

\subsection{Network model and safety verification}

Consider a feedforward neural network with $L$ hidden layers. Let $x \in \mathbb{R}^{n_x}$ denote the input, $\varphi^\ell,\vartheta^\ell \in \mathbb{R}^{n_\ell}$ denote the preactivation and postactivation vectors of layer $\ell$, and $y \in \mathbb{R}^{n_y}$ denote the output. We write the network as
\begin{align*}
\vartheta^0 &= x, \\
\varphi^\ell &= W^\ell \vartheta^{\ell-1} + b^\ell,
\qquad \ell=1,\dots,L, \\
\vartheta^\ell &= \boldsymbol{\sigma}(\varphi^\ell),
\qquad \ell=1,\dots,L, \\
y &= W^{L+1}\vartheta^L + b^{L+1},
\end{align*}
where $W^\ell \in \mathbb{R}^{n_\ell \times n_{\ell-1}}$, $b^\ell \in \mathbb{R}^{n_\ell}$ for $\ell=1,\dots,L$, and $W^{L+1}\in\mathbb{R}^{n_y \times n_L}$, $b^{L+1}\in\mathbb{R}^{n_y}$. The scalar activation is denoted by $\sigma:\mathbb{R}\to\mathbb{R}$, and $\boldsymbol{\sigma}$ is the corresponding componentwise map,
\begin{equation*}
\boldsymbol{\sigma}(v):=
\begin{bmatrix}
\sigma(v_1) & \cdots & \sigma(v_n)
\end{bmatrix}^{\top},
\qquad
v\in\mathbb{R}^n.
\end{equation*}
For clarity, the formulation is presented for a common activation function $\sigma$ across the hidden layers. The extension to layer- or neuron-dependent activations is immediate.

Let $\Xset:=\{x\in\mathbb{R}^{n_x} \, | \, [x \ \ 1] P_i [x \ \ 1]^\top\geq 0\ \text{for } i=1,\ldots,N_x\}$, for some $N_x\in\mathbb{Z}_+$ and $P_i\in\mathbb{S}^{n_x+1}$, be a bounded admissible input set and $\Yset$ its associated output reachable set, i.e., 
\begin{equation*}
\Yset:= \left\{y \in \mathbb{R}^{n_y} \, | \, y=f(x),\ x\in\Xset \right\},
\end{equation*}
where $f$ denotes the input-output map of the network and is bounded on $\Xset$. Define the function $b^\star:\mathbb{R}^{n_y}\rightarrow \mathbb{R}$ by
\begin{equation*}
b^\star(a):=\sup_{x\in\Xset}\, a^\top f(x).
\end{equation*}
Then, clearly, for any upper bound $\bar{b}_a\ge b^\star(a)$, the halfspace $\mathcal{H}_{\bar{b},a}:=\left\{y\in\mathbb{R}^{n_y} \mid a^\top y \le \bar{b}_a\right\}$ is a superset of $\Yset$, i.e., $\Yset \subseteq \mathcal{H}_{\bar{b},a}$.
Thus, given prescribed directions $a_i\in\mathbb{R}^{n_y}$, $i=1,\dots,N_{\mathrm{dir}}$ for some $N_{\mathrm{dir}}\in\mathbb{Z}_+$, a polytopic outer approximation is~given~by
\begin{equation*}
\Yset\subseteq  \left\{y\in\mathbb{R}^{n_y} \mid a_i^\top y \le \bar{b}_{a_i},\ i=1,\dots,N_{\mathrm{dir}} \right\}.
\end{equation*}
Methods such as the one in \cite{fazlyab2022} provide appropriate upper bounds $\bar{b}_{a_i}$ (the minimal values afforded by the method used). If the network's safety requirement can be expressed in terms of a set of linear inequalities, $a^\top_{\mathrm{safe},i}y\leq b_{\mathrm{safe},i}$, then verifying safety entails showing that the safety region defined by these inequalities is a superset of the aforementioned polytopic outer approximation, i.e., $\bar{b}_{a_{\mathrm{safe},i}}\leq b_{\mathrm{safe},i}$, where $\bar{b}_{a_{\mathrm{safe},i}}$ is obtained by applying a reachability analysis result. Alternatively, the safety conditions can be incorporated directly into the reachability analysis problem, and then safety is proved by solving a feasibility problem. The same framework also supports other safety and outer approximation set descriptions, such as ellipsoids~\cite{fazlyab2022}.

\subsection{QC-based SDP formulation}

Let $\mathbb{Q}_{\mathrm{cert}} \subset \mathbb{S}^3$ denote the set of symmetric matrices associated with the verified quadratic functions in $\Qcert$. The scalar activation $\sigma$ is said to satisfy the QCs defined by $\mathbb{Q}_{\mathrm{cert}}$ if, for every pair $(h_1,h_2)$ with $h_1 \in \Dset$ and $h_2 = \sigma(h_1)$,
\begin{equation*}
\begin{bmatrix}
h_1 & h_2 & 1
\end{bmatrix}
Q
\begin{bmatrix}
h_1 \\ h_2 \\ 1
\end{bmatrix}
\ge 0
\ \
 \mbox{for all}\  Q \in \mathbb{Q}_{\mathrm{cert}}.
\end{equation*}
If that is the case, then $\sigma$ also satisfies the QCs defined by the conic hull of $\mathbb{Q}_{\mathrm{cert}}$.

Similarly to \cite{fazlyab2022}, we define the lifted vector
\begin{equation*}
\xi :=
\begin{bmatrix}
x^\top & (\vartheta^1)^\top & \cdots & (\vartheta^L)^\top & 1
\end{bmatrix}^\top,
\end{equation*}
along with extraction matrices to recover the variables of interest from the lifted vector. Specifically, let $E_x\xi=[\,x \ \ 1\,]^\top$, $E_y\xi=[\,y\ \ 1\,]^\top$, and $E_i^\ell\xi=[\,\varphi_i^\ell\ \ \vartheta_i^\ell\ \ 1\,]^\top$ for each neuron $i$ in layer $\ell$.
These matrices are constructed based on the network weights, biases, and the definition of $\xi$. 

The linear matrix inequality from the reachability analysis SDP in \cite{fazlyab2022} follows from the S-procedure \cite{boyd1994linear} and has~the~form
\begin{equation}
M_x + M_{\boldsymbol{\sigma}} + M_y \preceq 0,\label{eq:lmi_reach}
\end{equation}
where the dependence of the first two terms on multipliers is suppressed for simplicity. Here, $M_x=E_x^\top \left(\sum_{i=1}^{N_x}\tau_iP_i\right) E_x$, with $\tau=[\tau_1\ \cdots \ \tau_{N_x}]^\top$ denoting the vector of nonnegative multipliers, corresponds to the input-set constraints, $M_{\boldsymbol{\sigma}}$ corresponds to the activation-related quadratic constraints, and $M_y$ corresponds to the conditions defining the overapproximated output set or the output safety constraints.
The admissible input set $\Xset$, the overapproximated output set, and the safety region are typically described by hyperrectangles, polytopes, or ellipsoids, all of which can be handled within this QC-based SDP framework \cite{fazlyab2022}. In the polytopic case, the normal vectors to the polytope facets are prescribed by the user. 

The verified quadratic characterizations derived in Section~\ref{sec:qc_construction} are incorporated into the framework via the matrix
\begin{equation*}
M_{\mathrm{cert}}(\Lambda)
=
\sum_{\ell=1}^{L}
\sum_{i=1}^{n_\ell}
\sum_{r=1}^{N_{\ell,i}}
\lambda_{\ell,i,r}\,
(E_i^\ell)^\top Q_{\ell,i,r} E_i^\ell,
\end{equation*}
where $\Lambda$ denotes the vector of the nonnegative multipliers $\lambda_{\ell,i,r}$ associated with the verified scalar quadratic constraints, and $\{Q_{\ell,i,r}\}_{r=1}^{N_{\ell,i}}$ denotes the family of symmetric matrices defining the verified scalar QCs assigned to neuron $i$ in layer $\ell$. In the common case where all neurons share the same activation function and the same verified family of QCs, we have $N_{\ell,i}=N_q$ and $Q_{\ell,i,r}=Q_r$ for all $\ell$ and $i$, so that the verified QC family is identical across the network, while the multipliers remain neuron/layer-dependent.

Multiple activation characterizations or properties can be combined within the overall activation block $M_{\boldsymbol{\sigma}}$. For example, local bounds on preactivations or postactivations can be encoded as quadratic inequalities and added to the formulation. Accordingly, we write
\begin{equation}\label{eq:Msigma}
M_{\boldsymbol{\sigma}}=M_{\mathrm{cert}}(\Lambda)+M_{\mathrm{loc}}(\nu)+\cdots,
\end{equation}
where $M_{\mathrm{loc}}(\nu)$, with $\nu$ denoting the vector of nonnegative multipliers, corresponds to the local-bound constraints, and the rest of the terms correspond to any other valid constraints.

The following proposition follows from the results of \cite{hu2020,fazlyab2022}. It addresses the computation of a polytopic overapproximation for the output reachable set.
\vskip 0.05in
\begin{prop}[\!\!\cite{hu2020,fazlyab2022}]
Let $a_j \in \mathbb{R}^{n_y}$ be a normal vector to facet $j$ of the overapproximating polytope, and define
\begin{equation*}
S(a_j,b_j)
:=
\begin{bmatrix}
0 & a_j \\
a_j^\top & -2b_j
\end{bmatrix}.
\end{equation*}
Then, the offset $b_j$ that, together with the normal vector $a_j$, defines the hyperplane $\{y\in\mathbb{R}^{n_y}\,|\, a_j^\top y= b_j\}$ containing the facet is obtained by solving the following optimization~problem:
\begin{equation}
\label{eq:polytope_reach}
\begin{aligned}
\min_{b_j,\Lambda,\nu,\tau,\ldots}\quad & b_j \\
\text{s.t.}\quad
& M_x + M_{\boldsymbol{\sigma}} + E_y^\top S(a_j,b_j) E_y \preceq 0, \\
&\Lambda\ge 0,\quad \nu \ge 0,\quad \tau \ge 0,\quad \ldots,
\end{aligned}
\end{equation}
where the inequality $\geq$ holds elementwise. Solving \eqref{eq:polytope_reach} over the selected family of facet normal vectors yields a polytopic overapproximation of the output reachable set.
\end{prop}
\vskip 0.05in
Clearly, pre- and post-multiplying the matrix inequality in \eqref{eq:polytope_reach} by $\xi^\top$ and $\xi$, respectively, and observing that both $\xi^\top M_x \xi$ and $\xi^\top M_{\boldsymbol{\sigma}} \xi$ are nonnegative for every admissible trajectory yields $\xi^\top E_y^\top S(a_j,b_j) E_y \xi \le 0$, or equivalently, $a_j^\top y \leq b_j$ holds for all $x \in \Xset$.
 
For safety verification, a single linear inequality $c^\top y \le d$ is checked by solving the SDP feasibility problem
\begin{equation}
\label{eq:safety_feas}
\begin{aligned}
\text{find}\quad & \Lambda,\nu,\tau\ldots \\
\text{s.t.}\quad
& M_x + M_{\boldsymbol{\sigma}} + E_y^\top S(c,d) E_y \preceq 0, \\
& \Lambda\ge 0,\quad \nu \ge 0,\quad \tau \ge 0, \quad \ldots
\end{aligned}
\end{equation}
If \eqref{eq:safety_feas} is feasible, the inequality $c^\top y \le d$ holds for all $x \in \Xset$. More general polyhedral safety constraints can be handled by iteratively applying this procedure to each defining inequality.

\begin{rmk}\label{rmk:scope}
The preceding construction is not restricted to feedforward neural networks. More generally, any analysis framework for uncertain or nonlinear systems that models a relation of the form $\vartheta=\Delta(\varphi)$ through pointwise IQCs can, in principle, use the verified quadratic characterizations developed in Section~\ref{sec:qc_construction}, provided the relation is represented on a suitable bounded operating domain. This includes, for example, QC-based procedures for reachability analysis, invariant-set computation, and safety verification. The same idea extends to formal-verification settings in which QCs serve as annotations or preconditions, such as the deductive verification of control programs in \cite{khalife2025nfm}.
\end{rmk}

\section{ReLU-Specific Tightening for Feedforward Neural Network Reachability Analysis}\label{sec:relu_tightening}

Section~\ref{sec:reachability_application} establishes how verified quadratic characterizations can be embedded in a QC-based SDP for reachability analysis of feedforward neural networks. For ReLU networks, however, the method of Section~\ref{sec:qc_construction} is not needed since the scalar ReLU graph already admits an exact quadratic description. In this case, the main sources of conservatism are the treatment of neurons as independent units and the overapproximation of local bounds. This section introduces ReLU-specific improvements based on exact scalar constraints, blockwise repeated-ReLU coupling, local bound tightening, and neuron pruning.

\subsection{ReLU characterizations}

The graph of the scalar ReLU relation, $\vartheta=\relu(\varphi)=\max{(0,\varphi)}$, is characterized exactly by
\begin{equation}
\vartheta \ge 0,\quad \vartheta \ge \varphi,\quad\vartheta\,(\varphi-\vartheta) \ge 0.
\label{eq:relu_exact_scalar}
\end{equation}
When imposed separately on each preactivation-postactivation pair, this exact scalar characterization treats the ReLU units independently and does not capture relations among neuron activations, which introduces conservatism in the SDP relaxation.
Repeated-nonlinearity QCs, as in \cite{fazlyab2022}, partially address this issue by correlating repeated copies of the same activation, thereby reducing conservatism at an increased computational cost. A more recent result in \cite{noori2025} derives a complete QC characterization for repeated ReLU in terms of matrix copositivity conditions. However, directly applying this complete characterization across all hidden-layer units remains computationally prohibitive.

To maintain tractability, we employ a relaxed blockwise construction. Let $\varphi_r,\vartheta_r\in\mathbb{R}^{n_r}$ consist of the preactivation and postactivation variables of the ReLU units included in the repeated characterization. Let $\mathcal{I}_1,\dots,\mathcal{I}_{N_b}$ form a partition of $\{1,\dots,n_r\}$ for some $N_b\in\mathbb{Z}_+$, with
\begin{equation*}
|\mathcal{I}_i| \le s_{\max},
\qquad
i=1,\dots,N_b,
\end{equation*}
where $s_{\max}$ is a user-specified maximum block size. From \cite{noori2025}, for each block $\mathcal{I}_i$, we impose a repeated-ReLU QC defined by a matrix of the form
\begin{equation}
M_{\mathrm{rep}}^{(i)} = M_1^{(i)} + (T^{(i)})^\top Q_2^{(i)} T^{(i)},
\label{eq:relu_block_M12}
\end{equation}
where
\begin{equation*}
M_1^{(i)}:=
\begin{bmatrix}
0 & Q_1^{(i)}\\
Q_1^{(i)} & -2Q_1^{(i)}
\end{bmatrix},
\qquad
T^{(i)}:=
\begin{bmatrix}
-I_{|\mathcal{I}_i|} & I_{|\mathcal{I}_i|}\\
0 & I_{|\mathcal{I}_i|}
\end{bmatrix},
\end{equation*}
and $Q_1^{(i)}$ is a diagonal matrix variable. The copositivity constraint on the matrix variable $Q_2^{(i)}$ is relaxed by introducing the variable $N_2^{(i)}$ and imposing
\begin{equation}
Q_2^{(i)} - N_2^{(i)} \succeq 0,
\qquad
N_2^{(i)} = (N_2^{(i)})^\top \ge 0,
\label{eq:relu_block_relax}
\end{equation}
where $N_2^{(i)} \ge 0$ holds elementwise. After pre- and post-multiplying \eqref{eq:relu_block_M12} by the appropriate extraction matrices, the resulting terms are added to the expression for $M_{\boldsymbol{\sigma}}$ in \eqref{eq:Msigma}. The blockwise repeated-ReLU characterization retains relations between neuron activations while keeping the SDP tractable. The blocks can be formed sequentially on the stacked index set and are therefore not restricted to a single layer. Alternatively, blocks may be formed within each layer by grouping neurons with similar incoming weight directions, e.g., by prioritizing pairs $(i,j)$ with high cosine similarity $\mathrm{sim}_{ij}^\ell := \frac{W_i^\ell (W_j^\ell)^\top}{\|W_i^\ell\|_2\,\|W_j^\ell\|_2}$, where $W_i^\ell$ denotes the $i$th row of $W^\ell$.

\subsection{Local bounds and pruning}

A second source of conservatism is the overapproximation of local neuron bounds. Let $\underline{\varphi}_i^\ell \le \varphi_i^\ell \le \overline{\varphi}_i^\ell$ denote local preactivation bounds for neuron $i$ in layer $\ell$. Such bounds may be computed inexpensively by interval bound propagation (IBP) and incorporated into the SDP through quadratic box constraints, for example,
\vspace{-0.75mm}\begin{equation*}
(\varphi_i^\ell-\underline{\varphi}_i^\ell)
(\overline{\varphi}_i^\ell-\varphi_i^\ell)\ge 0,
\quad
(\vartheta_i^\ell-\underline{\vartheta}_i^\ell)
(\overline{\vartheta}_i^\ell-\vartheta_i^\ell)\ge 0.\vspace{-0.75mm}
\end{equation*}

These local bounds also enable pruning. If $\overline{\varphi}_i^\ell \le 0$, then neuron $i$ is always inactive and $\vartheta_i^\ell=0$. Such a neuron is removed from the analysis by deleting the $i$th row of $W^\ell$, the $i$th entry of $b^\ell$, and the $i$th column of $W^{\ell+1}$. If $\underline{\varphi}_i^\ell \ge 0$, then neuron $i$ is always active and $\vartheta_i^\ell=\varphi_i^\ell$, i.e., the ReLU nonlinearity can be replaced with the identity relation. Only unstable neurons, i.e., neurons satisfying $\underline{\varphi}_i^\ell < 0 < \overline{\varphi}_i^\ell$, need to be characterized.
Pruning reduces the SDP dimension and avoids the conservatism of applying the ReLU characterization to neurons whose activation mode is already determined.

\subsection{Local bound tightening}

Although IBP is inexpensive, it suffers from the wrapping effect; as the depth increases, the propagated interval boxes become conservative because relations among neuron activations are ignored. In particular, IBP may admit combinations of neuron values that are individually feasible but not jointly realizable by the network. To mitigate this effect, we use a procedure for propagating bounding polytopes across the hidden layers. At each hidden layer $\ell$, we compute a postactivation polytope
$\mathcal{P}^\ell:=\left\{\vartheta^\ell\in\mathbb{R}^{n_\ell}\,|\, A^\ell \vartheta^\ell \le \bar{b}^\ell \right\}$, where the inequality is applied elementwise, the matrix $A^\ell$ is prescribed, and the vector $\bar{b}^\ell$ contains the tightest valid bounds, computed one bound at a time using the same QC-based SDP framework. The normal vectors to the polytope facets, i.e., the rows of $A^{\ell}$, may be selected progressively. At least two normal vectors are chosen for each neuron $i$, namely, $e_i$ and $-e_i$, where $e_i$ is the $i^{\mathrm{th}}$ standard basis vector. This minimal choice yields componentwise bounds. Tighter descriptions are obtained by adding normal vectors such as
\begin{enumerate}
\item the right-singular vectors corresponding to the $t$ largest singular values, for some $t$, of the next-layer weight matrix, which help capture directions most relevant for propagation through the subsequent linear map \cite{xue2013restructuring};
\item the vectors $\pm e_i \pm e_j$, which encode pairwise relations between neurons (i.e., octagonal constraints).
\end{enumerate}
Additional facet families may be included in the same manner. In general, a richer facet set yields a tighter propagated polytope at a higher computational cost.

Once $\mathcal{P}^\ell$ is available, it is used to tighten the next-layer preactivation bounds by solving linear programs. For each neuron $i$ in layer $\ell+1$, we solve for
\begin{align*}
\underline{\varphi}_{i,\mathrm{LP}}^{\ell+1}&=\min_{\vartheta^\ell\in\mathcal{P}^\ell}\left(W_i^{\ell+1}\vartheta^\ell+b_i^{\ell+1}\right),\\
\overline{\varphi}_{i,\mathrm{LP}}^{\ell+1}&=\max_{\vartheta^\ell\in\mathcal{P}^\ell}
\left(W_i^{\ell+1}\vartheta^\ell+b_i^{\ell+1}\right).
\end{align*}
The intersections of the obtained preactivation/postactivation intervals and the current intervals obtained from IBP generally yield improved bounds, which strengthen the local box constraints, enable additional pruning, and improve the polytope computed at the next layer. The procedure is repeated sequentially through the network.

In the polytope-propagation problems at the hidden layers, we use the exact scalar characterization together with the local bound constraints. At this stage, each additional normal vector already requires a separate SDP, and the added cost of block-repeated coupling is typically not justified by the resulting improvement in the propagated hidden-layer bounds. In the final output-reachability computation, however, the combined characterization is preferred: the exact scalar constraints enforce the exact graph of each ReLU unit, the block-repeated constraints capture relations among subsets of unstable neurons, and the propagated local bounds further restrict the feasible set. Among the tractable formulations considered, this yields the tightest output bounds.

\begin{rmk}\label{rmk:bnb}
In branch-and-bound implementations, the expensive SDP-based layerwise polytope propagation can be executed solely as root-level preprocessing; because child subdomains are subsets of the root domain, these initial bounds remain trivially valid for all descendant nodes.
\end{rmk}

\section{Examples}\label{sec:examples}
In this section, we apply the proposed methods to scalar nonlinearities, an uncertain-system example with saturation, and neural-network reachability and verification problems. All parameter values, sample sets, verification scripts, and benchmark networks are available in the \href{https://github.com/e-khalife/quadratic_characterizations_verification}{GitHub repository}.

\subsection{Quadratic characterizations of scalar nonlinearities}
We apply the proposed data-driven candidate-generation and verification procedure to two scalar relations. The resulting quadratic characterizations are shown in Fig.~\ref{fig:activation_characterizations}.

\subsubsection{Hyperbolic tangent}
Consider the set
\begin{equation*}
\Sset_{\tanh}
:=
\left\{
(\varphi,\vartheta)\in\mathbb{R}^2 \mid
\vartheta=\tanh(\varphi),\ \varphi\in[-20,20]
\right\}.
\end{equation*}
Since $\tanh$ is an odd function, candidate generation is carried out on four overlapping subregions in the left half of the domain, namely $[-4,-1.5]$, $[-2,0]$, $[-2,-0.7]$, and $[-1.2,-0.5]$, and the remaining candidates are obtained by symmetry. The first two subregions are used to generate upper-bounding candidates, and the latter two are used to generate lower-bounding candidates. For the data-driven step, we use $\rho=10^{-3},\, \lambda_{\mathrm{loc}}=10,\, \lambda_{\mathrm{g}}=1,\, \lambda_{\mathrm{ext}}=10,$ and $\bar{\gamma}=10^{-2},$ together with $5000$ global samples. In addition, we include the analytic global constraint $1-\vartheta^2\ge 0$, which is valid on the graph of $\tanh$ and therefore does not require SOS verification.
Since $\tanh$ is nonpolynomial, the data-driven candidates are verified using the procedure of Section~\ref{sec:qc_construction} for nonpolynomial relations. The domain is partitioned into the two tail intervals $[-20,-6]$ and $[6,20]$, on which $\tanh$ is approximated by a constant, and eight interior subintervals with breakpoints $0$, $\pm 1$, $\pm 2.5$, $\pm 4$, and $\pm 6$, on which it is approximated by degree-seven Chebyshev interpolants, with the error bounds obtained from the interpolation remainder~\cite{mason2002chebyshev} and the rounding of the coefficients. The candidates are verified on each relaxed verification set. The final verified family consists of the analytic global constraint and eight data-driven constraints.

\subsubsection{Unit saturation}
Consider the unit saturation function
\begin{equation*}
\vartheta=\sat(\varphi):=
\begin{cases}
-1, & \varphi\le -1,\\
\varphi, & -1\le \varphi \le 1,\\
1, & \varphi\ge 1.
\end{cases}
\end{equation*}
We consider its graph on the compact domain $\varphi\in[-5,5]$. Since the function is odd, candidate generation is again carried out on one half of the domain, on two overlapping subregions, $[-5,-1.2]$ and $[-1.5,-0.5]$, covering the saturated branch and the corner region, respectively, and the remaining candidates are obtained by symmetry. The data-driven quadratic program is solved with $\rho=10^{-3},\,\lambda_{\mathrm{loc}}=10,\,\lambda_{\mathrm{g}}=1,\,\lambda_{\mathrm{ext}}=5,$ and $\bar{\gamma}=10^{-3}$, together with $500$ global samples. For each subregion, the candidate is shaped using local samples together with uniformly generated exterior samples and selected target exterior points. Both subregions are used to generate upper-bounding candidates; their mirror images bound the graph from below. Together with the analytic constraint $1-\vartheta^2 \ge 0$, the final verified family consists of five constraints.

Unlike $\tanh$, the saturation function is piecewise affine and admits an exact semialgebraic description. Therefore, candidate generation and verification are carried out directly on the exact graph. The generated candidates are verified using the SOS feasibility problems and, independently, in Rocq~\cite{bertot2013interactive}. In addition, the characterizations of both relations are verified in Frama-C~\cite{cuoq2012framac} using the SMT solvers Z3~\cite{demoura2008z3} and Alt-Ergo~\cite{conchon2013}, a setting in which floating-point round-off errors can also be accounted for~\cite{khalife2023nfm,khalife2025nfm}.

\begin{figure}[htbp]
    \begin{minipage}{0.5\linewidth}
        \centering
        \includegraphics[width=\linewidth]{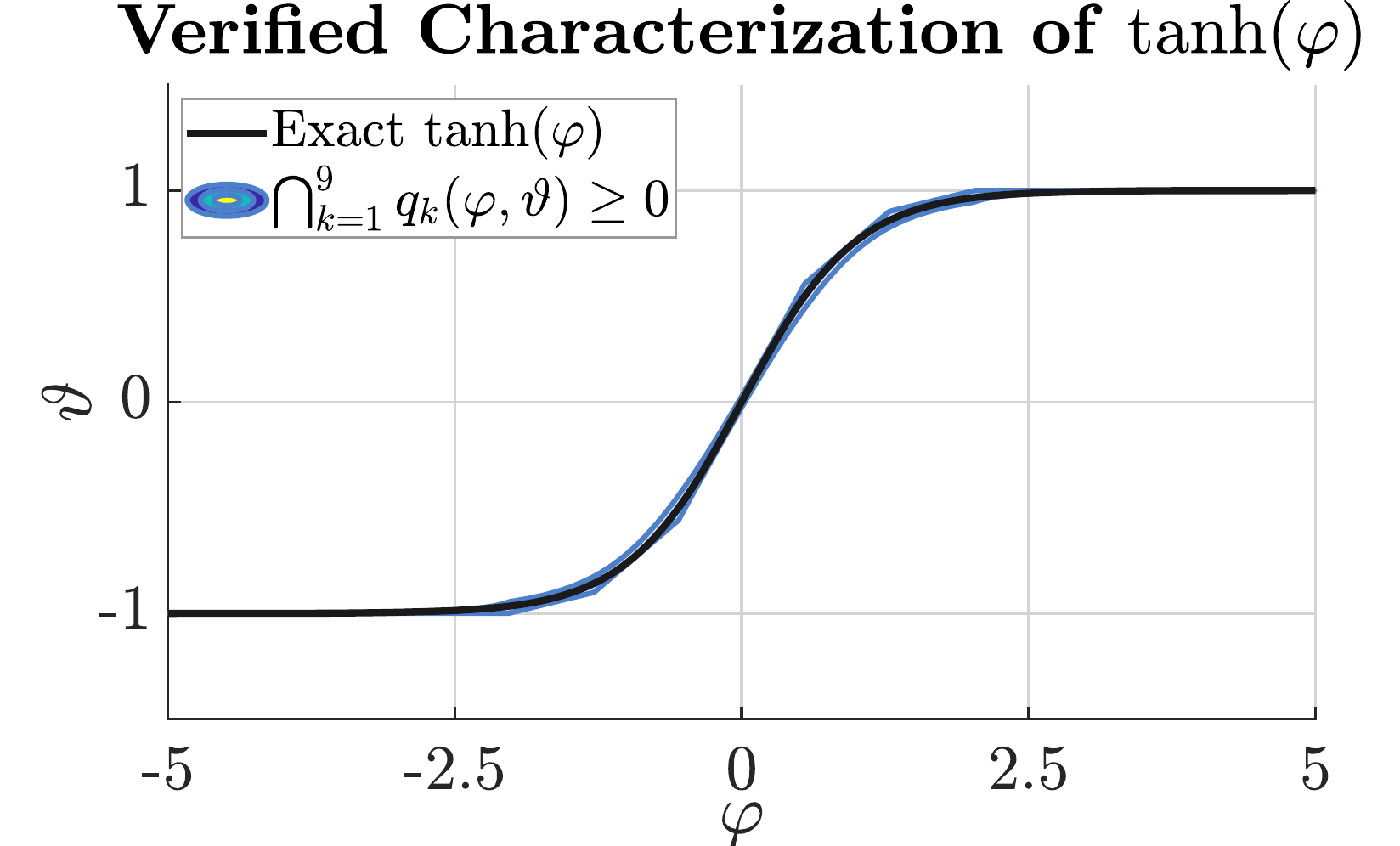}
    \end{minipage}
    \hfill
    \begin{minipage}{0.49\linewidth}
        \centering
        \includegraphics[width=\linewidth]{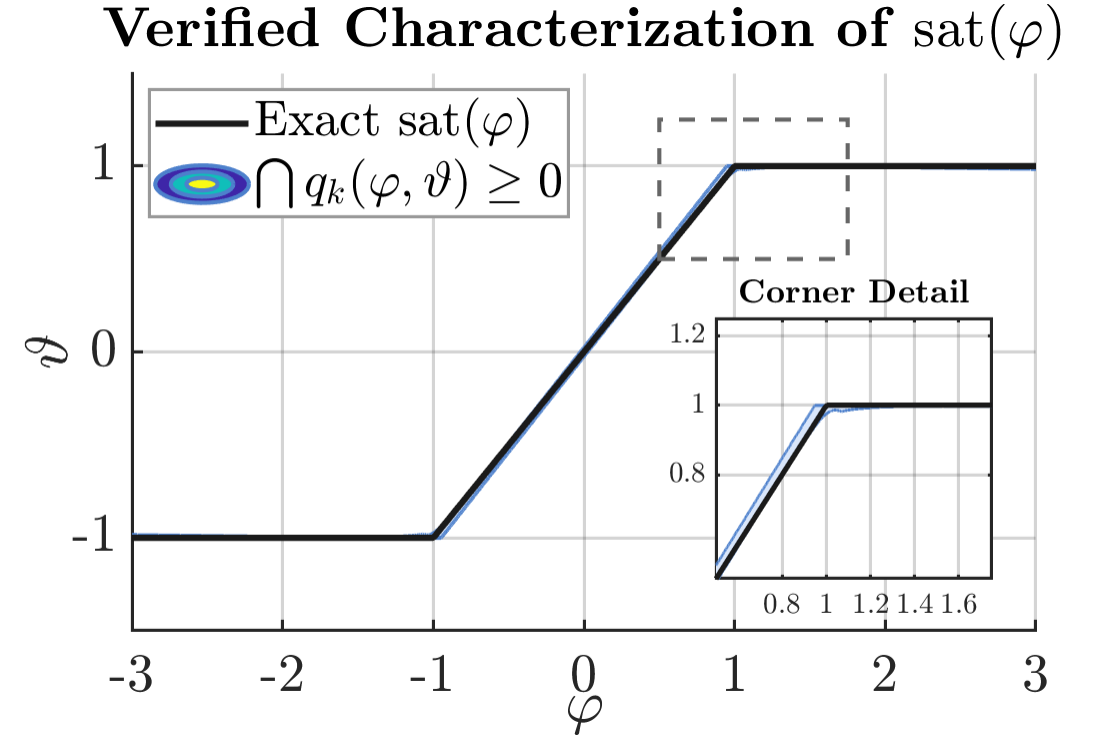}
    \end{minipage}
    \caption{Verified characterizations of $\tanh(\cdot)$ (left) and $\text{sat}(\cdot)$ (right).}
    \label{fig:activation_characterizations}
\end{figure}

\begin{figure}[htbp]
    \centering
    \includegraphics[width=\linewidth]{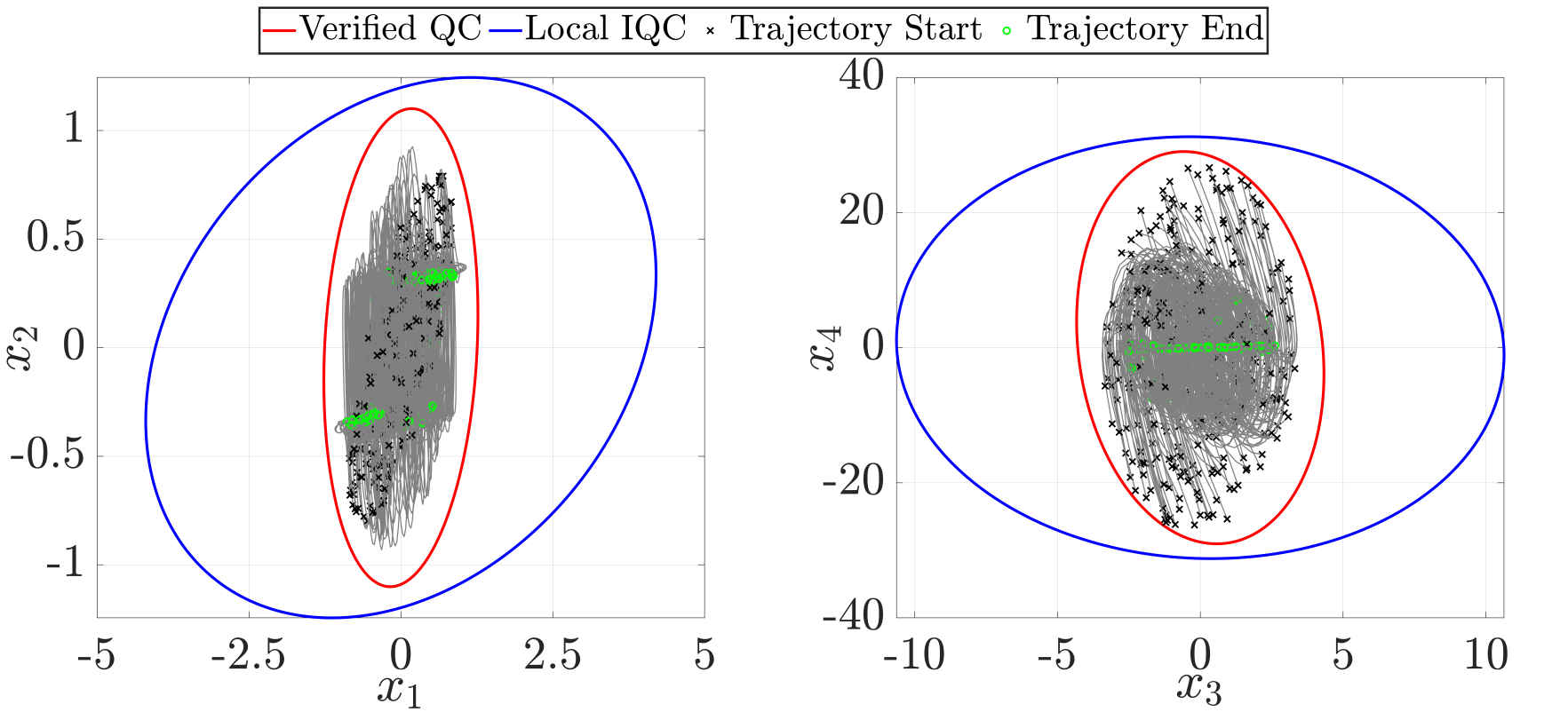}
    \caption{Projections of the state-invariant ellipsoids and simulations in the $(x_1,x_2)$ and $(x_3,x_4)$ planes.}
    \label{fig:invariant_ellipsoids_plot}\vspace{-3mm}
\end{figure}

To demonstrate the benefit of the proposed quadratic characterization for saturation, we consider the automobile active suspension example in \cite{khalife2023} with the same parameter values, controller synthesis, and uncertainty structure, consisting of an LTI uncertain parameter and a sector-bounded nonlinearity. The latter is defined as a saturation nonlinearity acting on the control input, with a saturation limit of $100\,\mathrm{N}$. Minimum-volume state-invariant ellipsoids are then computed using both the proposed characterization and the pointwise-IQC sector characterization of saturation with a locally tightened sector bound, following the methodology of~\cite{khalife2023}. In the pointwise-IQC approach, the sector characterization is enforced on the dead-zone representation of the saturation function, whereas the proposed characterization is imposed directly on the saturation input-output relation. Fig.~\ref{fig:invariant_ellipsoids_plot} shows the projections of the resulting invariant ellipsoids onto the $(x_1,x_2)$ and $(x_3,x_4)$ planes. The ellipsoid obtained with the proposed characterization is contained in the pointwise-IQC ellipsoid. The corresponding $-\log\det$ objective values are $30.95$ and $41.73$, respectively, for the augmented state ellipsoids, and $9.71$ and $14.60$ for the Schur-complement projections.

\subsection{Tanh Neural Networks Benchmark}\label{se:scalability_benchmark}
To evaluate the scalability and performance of the proposed method, we use the verified $\tanh$ characterization in Fig.~\ref{fig:activation_characterizations} within the QC-based SDP formulation of Section~\ref{sec:reachability_application} to benchmark against three reachability tools: NNV \cite{tran2020nnv}, CORA \cite{althoff2025manual}, and $\alpha$-CROWN \cite{xu2021fast}. The benchmark is conducted on feedforward networks with $\tanh$ activations, two outputs, and randomly generated weights. The input set is $\Xset = \{x \in \mathbb{R}^{n_x} \mid \|x\|_\infty \leq 1\}$. For every network, the preactivation bounds computed by IBP lie within $[-20, 20]$, the domain on which the $\tanh$ characterization is verified. The networks' architectures are varied over input dimension $n_x \in \{10, 20, 40\}$, number of hidden layers $L \in \{2, 4, 6\}$, and layer width $N \in \{16, 32, 64\}$. For each of the 27 configurations, the reported results are averages over 100 randomly generated networks.

The QC-based SDP is solved using the parser YALMIP \cite{Lofberg2004} with the solver MOSEK \cite{mosek}. NNV computes reachable sets via its \texttt{approx-star} method, which relies on star-set approximations, while CORA computes them using its zonotope propagation methods for nonlinear activations. For $\alpha$-CROWN, bounds are computed using the \texttt{auto\_LiRPA} \cite{xu2020automatic} library with 20 optimization iterations. The QC-based approach, NNV, and CORA are run in MATLAB on an Intel Core i9 Windows machine with 32\,GB of RAM, and $\alpha$-CROWN on Google Colab with GPU acceleration.

\paragraph{Tightness of Output Bounds}
For every output, each tool computes an upper and a lower bound, defining an output interval. To assess the tightness of these bounds, we compute the percentage difference between the output-interval width obtained from each tool and that of the QC-based method, averaged over the two outputs and the 100 networks. Positive values indicate a wider interval relative to the QC-based method, while negative values indicate a narrower interval.

Table~\ref{tab:tightness} reports these values for selected architecture configurations. To isolate the effect of the verified data-driven characterization, we also solve the same SDP with the generic QCs of $\tanh$~\cite{fazlyab2022}, namely, the sector constraint $\vartheta(\varphi-\vartheta) \geq 0$ and the bound $1-\vartheta^2 \geq 0$, instead of the verified family (column Gen. in Table~\ref{tab:tightness}). The generic characterization produces bounds between $38.7\%$ and $94.8\%$ wider than the verified one across all 27 configurations, with the relative width generally increasing with the number of hidden layers $L$. NNV produces bounds between $35.5\%$ and $103.8\%$ wider than the QC-based method across all 27 configurations, with the relative width generally increasing with the layer width $N$. CORA produces bounds between $10.0\%$ and $36.4\%$ wider, with the relative width also generally increasing with $N$. The relative tightness of $\alpha$-CROWN is highly dependent on the network architecture. For $N = 16$, its intervals are $7.9\%$ narrower than those of the QC-based method on average, with the maximum reduction of $24.5\%$ occurring at the largest tested depth ($L = 6$). For $N \ge 32$, the QC-based method yields tighter bounds in nearly all configurations, and at $N = 64$, it outperforms $\alpha$-CROWN across all tested depths and input dimensions, with the intervals of $\alpha$-CROWN up to $26.6\%$ wider.

Furthermore, as the input dimension $n_x$ increases, the relative width of $\alpha$-CROWN generally increases across most tested depths and widths. Consequently, the reduction in conservatism achieved by the proposed QC-based approach becomes more pronounced for networks with a larger number of inputs. This behavior is consistent with the loss of precision of linear relaxations in LiRPA when applied to wider networks \cite{xu2021fast}. These results indicate that the proposed characterization reduces the overapproximation error associated with a large number of neurons and high-dimensional inputs.

\begin{table}[t]
\centering
\caption{Average relative output-interval width (\%) of each tool w.r.t.\ the QC-based method for selected architecture configurations.}
\label{tab:tightness}
\begin{tabular}{ccc @{\hskip 1.5em} cccc}
\toprule
$n_x$ & $L$ & $N$ & Gen. & NNV & $\alpha$-CROWN & CORA \\
\midrule
\multirow{4}{*}{10}
& 2 & 16 & $+72.4$ & $+35.7$ & $-1.5$  & $+15.8$ \\
& 6 & 16 & $+94.7$ & $+57.2$ & $-24.5$ & $+10.7$ \\
& 2 & 64 & $+87.7$ & $+68.8$ & $+2.0$  & $+25.1$ \\
& 6 & 64 & $+87.7$ & $+82.8$ & $+10.1$ & $+22.4$ \\
\midrule
\multirow{4}{*}{40}
& 2 & 16 & $+38.7$ & $+35.5$ & $+2.6$  & $+15.7$ \\
& 6 & 16 & $+93.3$ & $+61.4$ & $-13.8$ & $+10.0$ \\
& 2 & 64 & $+73.0$ & $+65.4$ & $+18.7$ & $+22.7$ \\
& 6 & 64 & $+84.4$ & $+87.6$ & $+24.8$ & $+29.1$ \\
\bottomrule
\end{tabular}
\end{table}

\paragraph{Scalability}
We assess the computational scalability of the methods by analyzing their average execution times. The QC-based method solves two SDPs per output, one per bound; for all methods, the reported times cover both bounds. As expected, CORA is the least computationally expensive tool, requiring $0.13\,\mathrm{s}$ or less per network due to its structurally inexpensive zonotope propagation. The execution time of $\alpha$-CROWN remains relatively stable with respect to the input dimension and layer width. It mainly scales with network depth, requiring approximately $3.3\,\mathrm{s}$ for networks with $L=2$ and up to $23.2\,\mathrm{s}$ for $L=6$.
For the proposed QC-based approach, the execution time scales moderately with the input dimension $n_x$. As detailed in Table~\ref{tab:timing}, the primary computational bottleneck for the QC-based method is the total number of neurons ($L\times N$), reflecting the growth in the corresponding SDP matrix dimension. For the largest architecture tested ($n_x=40$, $L=6$, $N=64$), the proposed method requires approximately $378.0\,\mathrm{s}$ on average. With two multipliers per neuron instead of nine, the generic characterization requires between $40\%$ and $95\%$ less computation time than the verified one across the 27 configurations.

\begin{table}[t]
\centering
\caption{Average total computation time (s) for selected architecture configurations.}
\label{tab:timing}
\setlength{\tabcolsep}{4pt}
\begin{tabular}{ccc @{\hskip 1.5em} ccccc}
\toprule
$n_x$ & $L$ & $N$ & QC & Gen. & NNV & $\alpha$-CROWN & CORA \\
\midrule
\multirow{4}{*}{10}
& 2 & 16 & $0.59$  & $0.36$ & $0.38$ & $3.35$  & $0.013$ \\
& 6 & 16 & $3.66$  & $0.65$ & $1.24$ & $23.16$ & $0.026$ \\
& 2 & 64 & $10.82$ & $1.28$ & $1.38$ & $4.03$  & $0.032$ \\
& 6 & 64 & $300.8$ & $14.6$ & $18.1$ & $22.6$  & $0.108$ \\
\midrule
\multirow{4}{*}{40}
& 2 & 16 & $1.65$  & $0.43$ & $0.34$ & $3.80$  & $0.012$ \\
& 6 & 16 & $6.29$  & $0.84$ & $1.55$ & $22.44$ & $0.028$ \\
& 2 & 64 & $17.67$ & $2.55$ & $1.79$ & $3.29$  & $0.040$ \\
& 6 & 64 & $378.0$ & $17.6$ & $22.3$ & $21.7$  & $0.123$ \\
\bottomrule
\end{tabular}
\end{table}

\subsection{Reachability Analysis of a tanh Neural-Network Controller}

\begin{figure}[htbp]
    \centering
    \includegraphics[width=\linewidth]{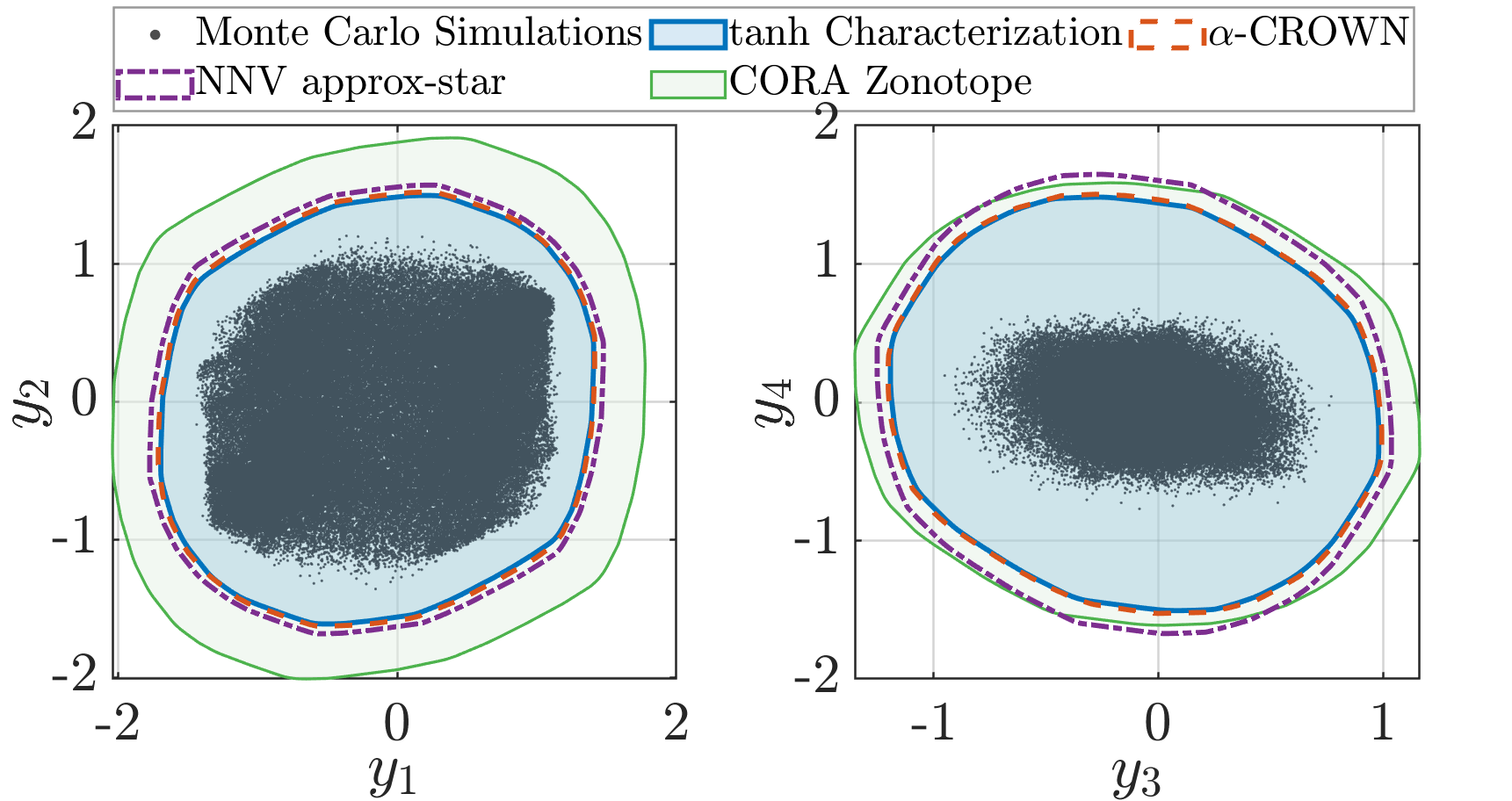}
    \vspace{-7.5mm}\caption{Output reachable sets in the $(y_1,y_2)$ and $(y_3,y_4)$ planes.}
    \label{fig:nn_reachable_sets}
\end{figure}

We consider the $\tanh$ neural-network controller from \cite{marquis2025adversarial}, with $27$ inputs, two hidden layers of $64$ neurons each, and $4$ outputs. The inputs are normalized measurements, so the admissible input set is $\Xset = \left\{x\in\mathbb{R}^{27}\mid \|x\|_\infty \le 1\right\}$.
The network outputs are control actions, which are post-processed before being applied to the system.
For this network, the preactivation bounds computed by IBP lie within $[-20,20]$, so the verified $\tanh$ characterization in Fig.~\ref{fig:activation_characterizations} can be used within the QC-based SDP formulation of Section~\ref{sec:reachability_application}. We compute output-reachability bounds and compare the resulting overapproximations against NNV, CORA, and $\alpha$-CROWN.
For the QC-based approach, NNV, and $\alpha$-CROWN, we construct the reachable-set overapproximation in the $(y_1,y_2)$ and $(y_3,y_4)$ planes by defining $180$ facet normal vectors per plane and computing the corresponding output bounds. In contrast, CORA computes the overapproximated reachable set via zonotope propagation. The resulting sets are plotted in Fig.~\ref{fig:nn_reachable_sets}.

Our proposed approach requires solving an SDP for each facet direction, taking approximately 0.08\,s per direction, or 28.5\,s in total. In contrast, CORA, NNV, and $\alpha$-CROWN require approximately 0.19\,s, 18.6\,s, and 3.6\,s, respectively, to compute their output bounds.  On average, over all facet directions, the bounds obtained with the proposed characterization are approximately $6.4\%$ tighter than those of NNV and approximately $1.2\%$ tighter than those of $\alpha$-CROWN in this example. The bounds obtained with the proposed method were also independently verified using $\alpha,\beta$-CROWN \cite{wang2021beta}. This illustrates that the adopted approach with the proposed characterization yields tighter single-pass bounds than comparable reachable-set overapproximation tools.

\subsection{ReLU Reachability and Verification on ACAS Xu}
We apply the ReLU-specific tightening methods of Section~\ref{sec:relu_tightening} to the ACAS Xu benchmark \cite{julian2016policy,katz2017reluplex}, a standard neural-network verification benchmark for airborne collision avoidance.
For simplicity, we refer to the joint use of the exact polynomial characterization of ReLU in \eqref{eq:relu_exact_scalar} and the blockwise repeated-ReLU characterization defined by \eqref{eq:relu_block_M12}--\eqref{eq:relu_block_relax} as the combined characterization.
For the reachability experiments, we consider the input set corresponding to Property $\phi_1$ from \cite{katz2017reluplex} and compare the exact polynomial characterization with IBP local bounds (EP), the combined characterization with IBP local bounds (COMB), and the combined characterization with polytope propagation (COMB-PP) against NNV, CORA, and $\alpha$-CROWN. In the local-bound tightening step, the normal vectors consist of the standard basis vectors, together with $25$ right-singular vectors, and the repeated-ReLU maximum block size, $s_{\max}$, is set to $10$.

Table~\ref{tab:acasxu_widths} reports, for each output, the average interval width over the ACAS~Xu networks. Across every output, COMB yields strictly tighter bounds than EP, and COMB-PP further tightens those bounds, achieving the smallest average widths of all methods considered. As detailed in the table, NNV and $\alpha$-CROWN produce more conservative bounds, while CORA yields the widest output-intervals by a large margin.
The effect of polytope propagation becomes more pronounced in deeper layers: the average local-width reduction relative to IBP is $39.3\%$ at the fifth hidden layer and $61.9\%$ at the sixth hidden layer, which leads to tighter output bounds in COMB-PP compared to COMB. Stronger tightening can be obtained by using more normal vectors, such as ones encoding pairwise relations between neurons, or by increasing the repeated-ReLU block size, though both lead to higher computational cost. In particular, with the aforementioned normal vectors added, the deepest-layer average local-width reduction reaches approximately $96$--$98\%$ relative to IBP in our experiments.

\begin{table}[t]
\centering
\caption{Average output-interval widths for ACAS~Xu Property~$\phi_1$ (ratio to COMB-PP in parentheses).}
\label{tab:acasxu_widths}
\scriptsize 
\setlength{\tabcolsep}{2.5pt}
\renewcommand{\arraystretch}{0.95}
\begin{tabular}{lrrrrr}
\toprule
Method & $y_1$ & $y_2$ & $y_3$ & $y_4$ & $y_5$ \\
\midrule
\textbf{COMB-PP} & \textbf{93.1} & \textbf{75.5} & \textbf{77.4} & \textbf{89.2} & \textbf{87.0} \\
COMB    & 128.2\,(1.4) & 107.9\,(1.4) & 110.9\,(1.4) & 126.4\,(1.4) & 120.5\,(1.4) \\
EP      & 182.2\,(2.0) & 150.0\,(2.0) & 156.3\,(2.0) & 179.5\,(2.0) & 172.2\,(2.0) \\
NNV     & 332.0\,(3.6) & 255.9\,(3.4) & 268.4\,(3.5) & 308.2\,(3.5) & 305.5\,(3.5) \\
$\alpha$-CROWN & 373.3\,(4.0) & 288.4\,(3.8) & 302.5\,(3.9) & 347.2\,(3.9) & 344.1\,(4.0) \\
CORA    & 5241.5\,(56) & 4454.8\,(59) & 4657.4\,(60) & 4963.8\,(56) & 4799.3\,(55) \\
\bottomrule
\end{tabular}
\end{table}

Regarding computation time, NNV, $\alpha$-CROWN, and CORA require an average of 5.9\,s, 3.1\,s, and 0.16\,s per network, respectively. In contrast, with YALMIP and MOSEK, the average solver time per output bound is approximately 8.3\,s for EP, 239\,s for COMB, and 258\,s for COMB-PP (excluding the time required for the polytope propagation step). In the layerwise polytope propagation step of COMB-PP, the average solver time per direction increases from 0.03\,s at layer~1 to 14.2\,s at layer~6. Because the optimization problems for each direction within each layer are independent, they are dispatched in parallel using MATLAB's Parallel Computing Toolbox \texttt{parfor}.

Although the proposed approaches are computationally intensive, this offline cost is justified for safety-critical applications. While complete verifiers such as $\alpha,\beta$-CROWN can use branch-and-bound algorithms to find tighter or even exact reachable sets given enough time, our approaches compute sound bounds in a single pass. Additionally, the neuron bounds obtained from the polytope propagation can be provided to $\alpha$-CROWN, which reduces its resulting output-interval widths by approximately 33\% in our experiments.

Furthermore, using the proposed framework for property verification, we verified Properties~$\phi_1$, $\phi_3$, and $\phi_4$ of the ACAS Xu benchmark~\cite[Appendix VI]{katz2017reluplex} for all networks to which they apply. Property~$\phi_1$ requires that $y_1 \le 1500$ over the prescribed input set. Properties~$\phi_3$ and $\phi_4$ require that $y_1$ is not minimal, i.e., that $y_{i+1} \le y_1$ for at least one $i \in \{1,\ldots,4\}$. This amounts to certifying the disjunctive condition $\bigvee_{i=1}^4 \left(y_{i+1}\le y_1\right)$, or equivalently, that there exists at least one $i\in\{1,\ldots,4\}$ such that $y_{i+1}-y_1\le 0$. In a branch-and-bound framework, this means that verification on a branch is complete once at least one of these inequalities is certified. This can be done in two ways: either by solving the feasibility problem associated with each condition separately, or by solving a single feasibility problem in which the term $M_y$ in \eqref{eq:lmi_reach} is defined by
\begin{equation*}
    \xi^\top M_y \xi = \sum_{i=1}^4 \lambda_i \left(y_{i+1}-y_1\right),
\end{equation*}
where $\lambda_i$ are nonnegative multipliers; if this problem is feasible with at least one nonzero multiplier, the property holds on the input set considered. When it is infeasible on the full input set, the input set is split recursively, and the property is verified once the problem is feasible on every branch (see Remark~\ref{rmk:bnb}).

\section{Conclusion}\label{sec:conclusion}
This paper presents a framework for constructing verified quadratic characterizations of scalar static relations on bounded domains. The resulting verified quadratic constraints are directly compatible with existing QC- and pointwise-IQC-based analysis methods. In the examples, they yield less conservative results than the corresponding generic sector-based constraints; for $\tanh$, this comes at a higher computational cost. For ReLU networks, the proposed tightening yields the tightest bounds among the methods considered on the ACAS Xu benchmark. Future work will explore applications to broader classes of neural networks and uncertain nonlinear systems and the extension of the verification process to account for floating-point errors.

\bibliographystyle{IEEEtran}
\bibliography{IEEEfull,References}

\end{document}